\documentclass[10pt,twocolumn,letterpaper]{article}

\usepackage{cvpr}
\usepackage[utf8]{inputenc}
\usepackage{times}
\usepackage{epsfig}
\usepackage{graphicx}
\usepackage{amsmath}
\usepackage{amssymb}
\usepackage{textcomp}
\usepackage{enumitem}
\usepackage{microtype}
\usepackage{booktabs}
\usepackage{makecell}
\usepackage{color}
 \usepackage{array,multirow}

\usepackage[pagebackref=true,breaklinks=true,letterpaper=true,colorlinks,bookmarks=false]{hyperref}

\cvprfinalcopy 

\def\cvprPaperID{4701} 

\ifcvprfinal\fi
\begin{document}

\title{Fast Spatially-Varying Indoor Lighting Estimation}

\author{
    Mathieu Garon$^{ \star}$\thanks{Parts of this work were completed while Mathieu Garon was an intern at Adobe Research.} , \,\,
    Kalyan Sunkavalli$^\dagger$, \,\,
    Sunil Hadap$^\dagger$, \,\,
    Nathan Carr$^\dagger$, \,\,
    Jean-François Lalonde$^\star$ \vspace{3pt}\\
    $^\star$Université Laval, $^\dagger$Adobe Research\\
    {\tt\small \ mathieu.garon.2@ulaval.ca}
    \quad \tt\small \{sunkaval, hadap, ncarr\}@adobe.com
    \quad \tt\small \ jflalonde@gel.ulaval.ca\\
}
\maketitle

\pagenumbering{gobble}
\begin{abstract}
We propose a real-time method to estimate spatially-varying indoor lighting from a single RGB image. Given an image and a 2D location in that image, our CNN estimates a 5th order spherical harmonic representation of the lighting at the given location in less than 20ms on a laptop mobile graphics card. While existing approaches estimate a single, global lighting representation or require depth as input, our method reasons about local lighting without requiring any geometry information. We demonstrate, through quantitative experiments including a user study, that our results achieve lower lighting estimation errors and are preferred by users over the state-of-the-art. Our approach can be used directly for augmented reality applications, where a virtual object is relit realistically at any position in the scene in real-time.
\end{abstract}

\section{Introduction}

Estimating the illumination conditions of a scene is a challenging problem. An image is formed by conflating the effects of lighting with those of scene geometry, surface reflectance, and camera properties. Inverting this image formation process to recover lighting (or any of these other intrinsic properties) is severely underconstrained. Typical solutions to this problem rely on inserting an object (a light probe) with known geometry and/or reflectance properties in the scene (a shiny sphere~\cite{Debevec-siggraph-98}, or 3D objects of known geometry~\cite{georgoulis-iccv-17,weber-3dv-18}). Unfortunately, having to insert a known object in the scene is limiting and thus not easily amenable to practical applications.

Previous work has tackled this problem by using additional information such as depth~\cite{barron2013rgbd, maier2017intrinsic3d}, multiple images acquired by scanning a scene~\cite{gruber2012real, v.20181249, zhang-siga-16, zhang2018discovering} or user input~\cite{karsch-sig-11}. However, such information is cumbersome to acquire. Recent work~\cite{gardner-sigasia-17} has proposed a learning approach that bypasses the need for additional information by predicting lighting directly from a single image in an end-to-end manner. While~\cite{gardner-sigasia-17} represents a practical improvement over previous approaches, we argue that this technique is still not amenable for use in more interactive scenarios, such as augmented reality (AR). First, it cannot be executed in real-time since it decodes full environment maps. Second, and perhaps more importantly, this approach produces a \emph{single} lighting estimate for an image (more or less in the center of the image). However, indoor lighting is \emph{spatially-varying}: light sources are in close proximity to the scene, thus creating significantly different lighting conditions across the scene due to occlusions and non-uniform
light distributions.

\begin{figure}[!t]
\centering
\setlength{\tabcolsep}{1pt}
\begin{tabular}{cc}
\includegraphics[clip,width=.49\linewidth]{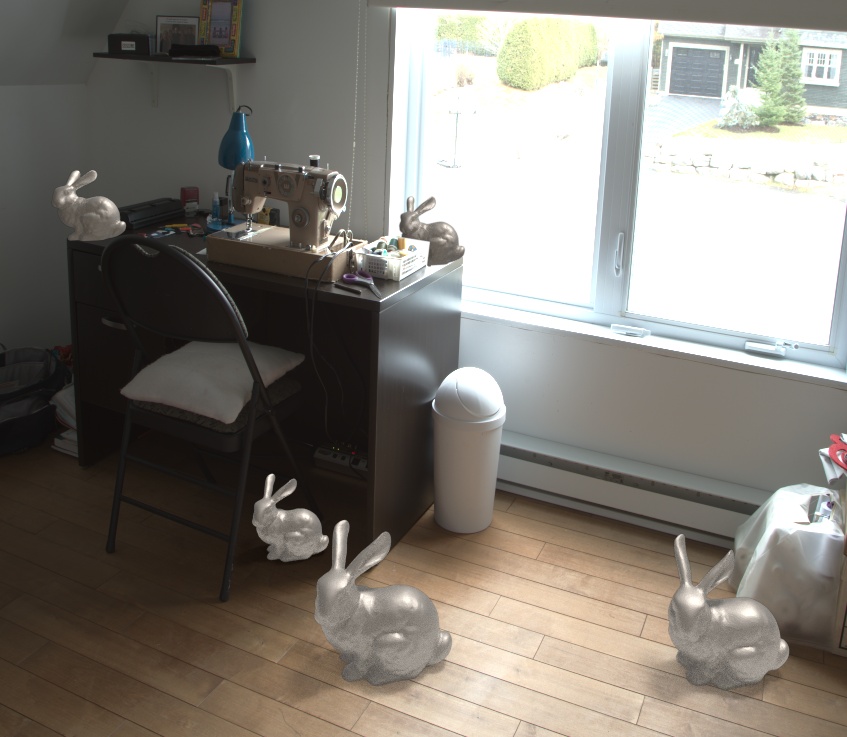} & 
\includegraphics[clip,width=.49\linewidth]{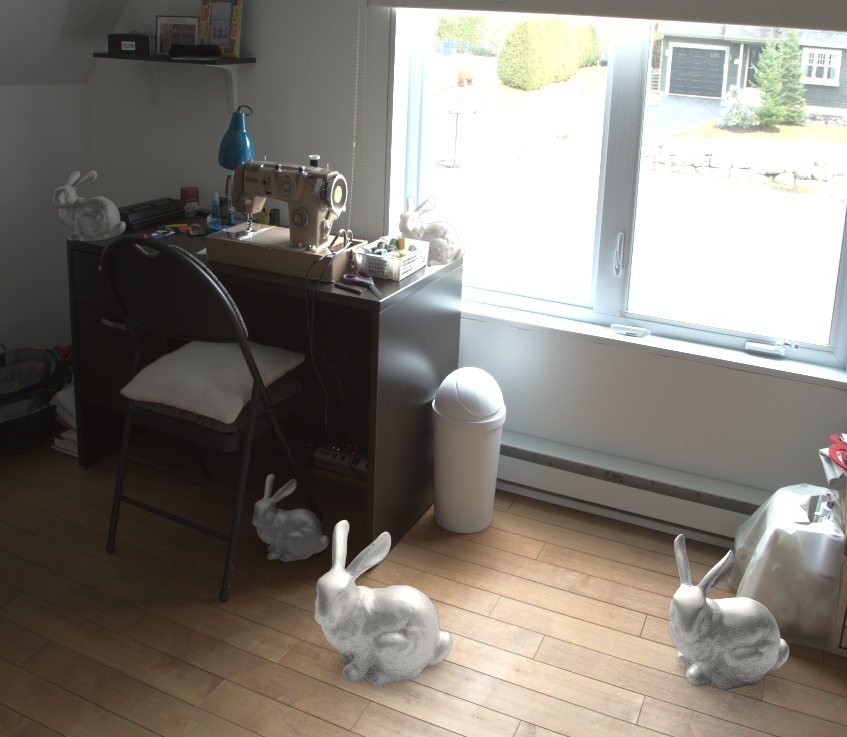} \\
\end{tabular}
\caption{Indoor lighting is spatially-varying. Methods that estimate global lighting~\cite{gardner-sigasia-17} (left) do not account for local lighting effects resulting in inconsistent renders when lighting virtual objects. In contrast, our method (right) produces spatially-varying lighting from a single RGB image, resulting in much more realistic results.}
\label{fig:teaser}
\end{figure}

Commercial AR systems, such as Apple's ARkit, provide real-time lighting estimation capabilities. However, we observe that their capabilities are quite rudimentary: ARkit uses ``exposure information to determine the relative brightness''\footnote{Source: WWDC 2017, session 602, \url{https://developer.apple.com/videos/play/wwdc2017/602/?time=2230}}, and requires the user to scan part of the environment and ``automatically completes [this map] with advanced machine learning algorithms''\footnote{Source: WWDC 2018, session 602, \url{https://developer.apple.com/videos/play/wwdc2018/602/?time=1614}}, but that is used only for environment mapping (realistic reflections) and not as a light source.

In this work, we present a method that estimates spatially-varying lighting---represented as spherical harmonics (SH)---from a \emph{single} image in real-time. Our method, based on deep learning, takes as input a single image and a 2D location in that image, and outputs the 5th-order SH coefficients for the lighting at that location. Our approach has three main advantages. First, spherical harmonics are a low-dimensional lighting representation (36 values for 5th-degree SH for each color channel), and can be predicted with a compact decoder architecture. Indeed, our experiments demonstrate that our network can predict 5th-degree SH coefficients in less than 20ms on a mobile GPU (Nvidia GTX970M). Second, the SH coefficients can directly be used by off-the-shelf shaders to achieve real-time relighting~\cite{ramamoorthi2001efficient,sloan2002precomputed}. Third, and perhaps more importantly, these local SH estimates directly embed local light visibility without the need for explicit geometry estimates. Our method therefore adapts to local occlusions and reflections without having to conduct an explicit reasoning on scene geometry. Note that while using SH constrains the \emph{angular} frequency of the lighting we can represent, by having a different estimate for every scene location, our method does capture high-frequency \emph{spatial} variations such as the shadowing under the desk in Figure~\ref{fig:teaser}(b).

To the best of our knowledge, our paper is the first to propose a practical approach for estimating spatially-varying lighting from a single indoor RGB image. Our approach enables a complete image-to-render augmented reality pipeline that automatically adapts to both local and global lighting changes at real-time framerates.
In order to evaluate spatially-varying methods quantitatively, a novel, challenging dataset containing 79 ground truth HDR light probes in a variety of indoor scenes is made publicly available\footnote{\scriptsize \url{https://lvsn.github.io/fastindoorlight/}}.
 
\section{Related work}

Estimating lighting from images has a long history in computer vision and graphics. In his pioneering work, Debevec proposed to explicitly capture HDR lighting from a reflective metallic sphere inserted into the scene~\cite{Debevec-siggraph-98}. 

A large body of work relies on more generic 3D objects present in the scene to estimate illumination. Notably, Barron and Malik~\cite{barron2015shape} model object appearance as a combination of its shape, lighting, and material properties, and attempt to jointly estimate all of these properties in an inverse rendering framework which relies on data-driven priors to compensate for the lack of information available in a single image. Similarly, Lombardi and Nishino~\cite{lombardi2016reflectance} estimate reflectance and illumination from an object of known shape. Prior work has also used faces in images to estimate lighting~\cite{blanz19993dmm}.

More recently, Georgoulis et al.~\cite{georgoulis2018reflectance} use deep learning to estimate lighting and reflectance from an object of known geometry, by first estimating its reflectance map (i.e., its ``orientation-dependent'' appearance)~\cite{rematas2016deep} and factoring it into lighting and material properties~\cite{georgoulis-iccv-17} afterwards.

Within the context of AR, real-time approaches \cite{gruber2012real,v.20181249} model the Radiance Transfer Function of an entire scene from its captured geometry, but require that the scene be reconstructed first. Zhang et al.~\cite{zhang-siga-16} also recover spatially-varying illumination, but require a complete scene reconstruction and user-annotated light positions. In more recent work, the same authors~\cite{zhang2018discovering} present a scene transform that identifies likely isotropic point light positions from the shading they create on flat surfaces, acquired using similar scene scans. Other methods decompose an RGBD image into its intrinsic scene components~\cite{barron2013rgbd, maier2017intrinsic3d} including spatially-varying SH-based lighting. In contrast, our work does not require any knowledge of scene geometry, and automatically estimates spatially-varying SH lighting from a single image. 

Methods which estimate lighting from single images have also been proposed. Khan et al.~\cite{khan-siggraph-06} propose to flip an HDR input image to approximate the out-of-view illumination. Similar ideas have been used for compositing~\cite{lalonde-sig-07}. While these approximations might work in the case of mostly diffuse lighting, directional lighting cannot be estimated reliably, for example, if the dominant light is outside the field of view of the camera. Karsch et al.~\cite{karsch-sig-11} develop a user-guided system, yielding high quality lighting estimates. In subsequent work~\cite{karsch-tog-14}, the same authors propose an automatic method that matches the background image with a large database of LDR panoramas, and optimizes light positions and intensities in an inverse rendering framework that relies on an intrinsic decomposition of the scene and automatic depth estimation. Lalonde et al.~\cite{lalonde-ijcv-12} estimate lighting by combining cues such as shadows and shading from an image in a probabilistic framework, but their approach is targeted towards outdoor scenes. Hold-Geoffroy et al.~\cite{holdgeoffroy-cvpr-17} recently provided a more robust method of doing so by relying on deep neural networks trained to estimate the parameters of an outdoor illumination model. 

More closely related to our work, Gardner et al.~\cite{gardner-sigasia-17} estimate indoor illumination from a single image using a neural network. Their approach is first trained on images extracted from a large set of LDR panoramas in which light sources are detected. Then, they fine-tune their network on HDR panoramas. Similarly, Cheng et al.~\cite{Cheng2018} proposed estimating SH coefficients given two images captured from the front and back camera of a mobile device. However, both of these works provide a single, global lighting estimate for an entire image. In contrast, we predict spatially-varying lighting.

\section{Dataset}
\label{sec:dataset}

\begin{figure}
\centering
\includegraphics[width=\linewidth]{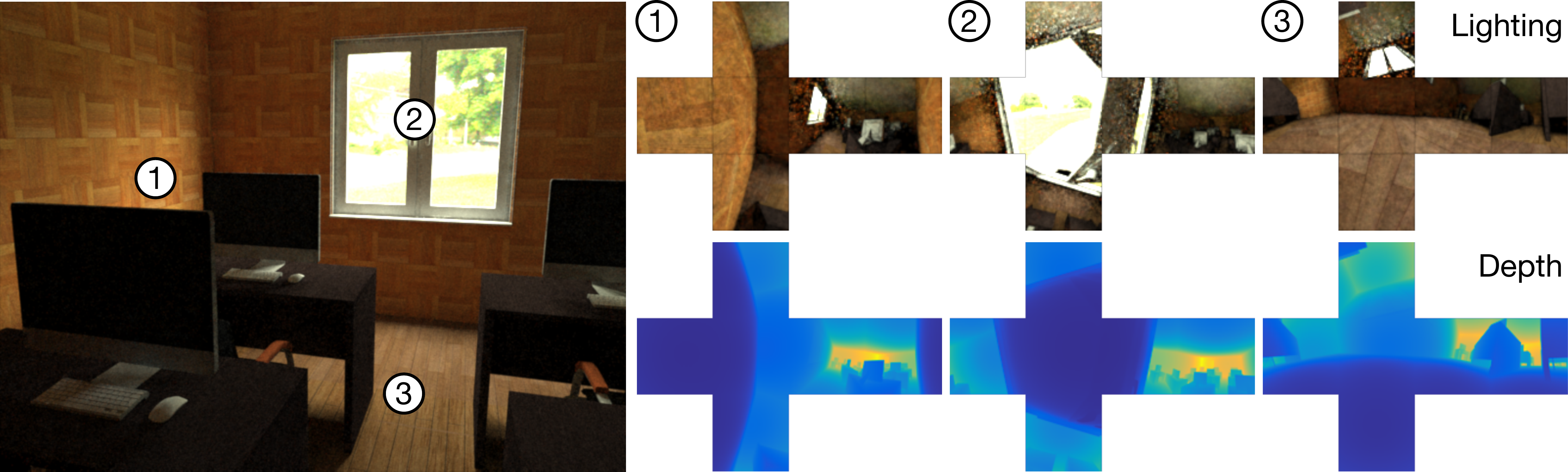}
\includegraphics[width=\linewidth]{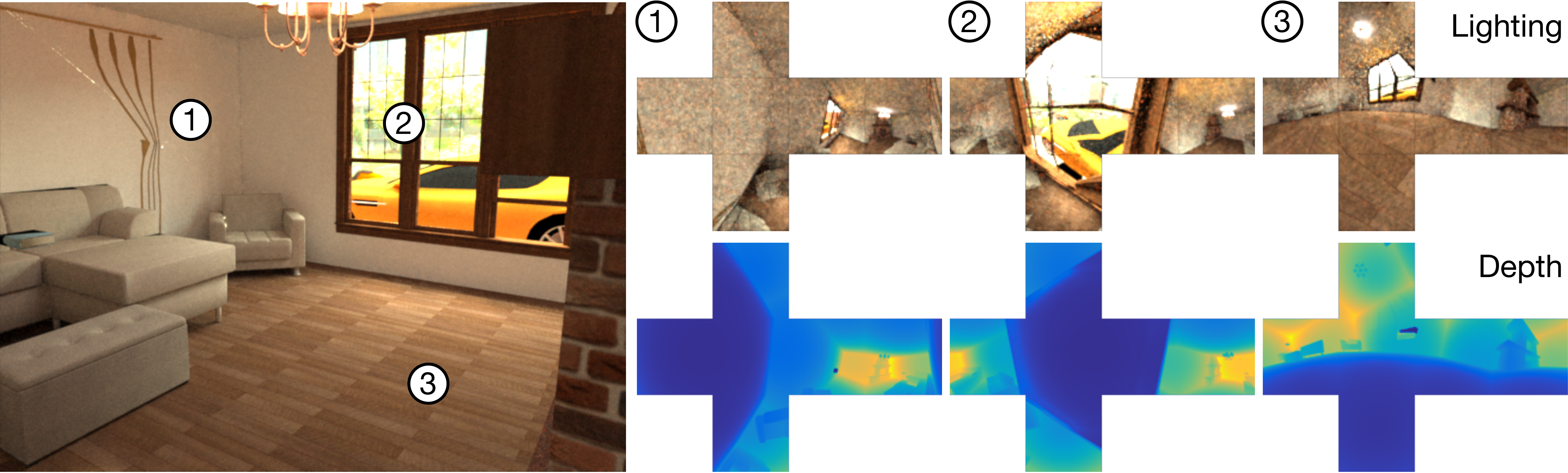}
\caption{Example synthetic light probes sampled in our dataset. Locations on the image are randomly sampled (left). For each location, light probes (right, top) and their corresponding depth maps (right, bottom) are rendered into cube maps.}
\label{fig:dataset}
\end{figure}

\begin{figure*}
\centering
\footnotesize
\includegraphics[width=\linewidth]{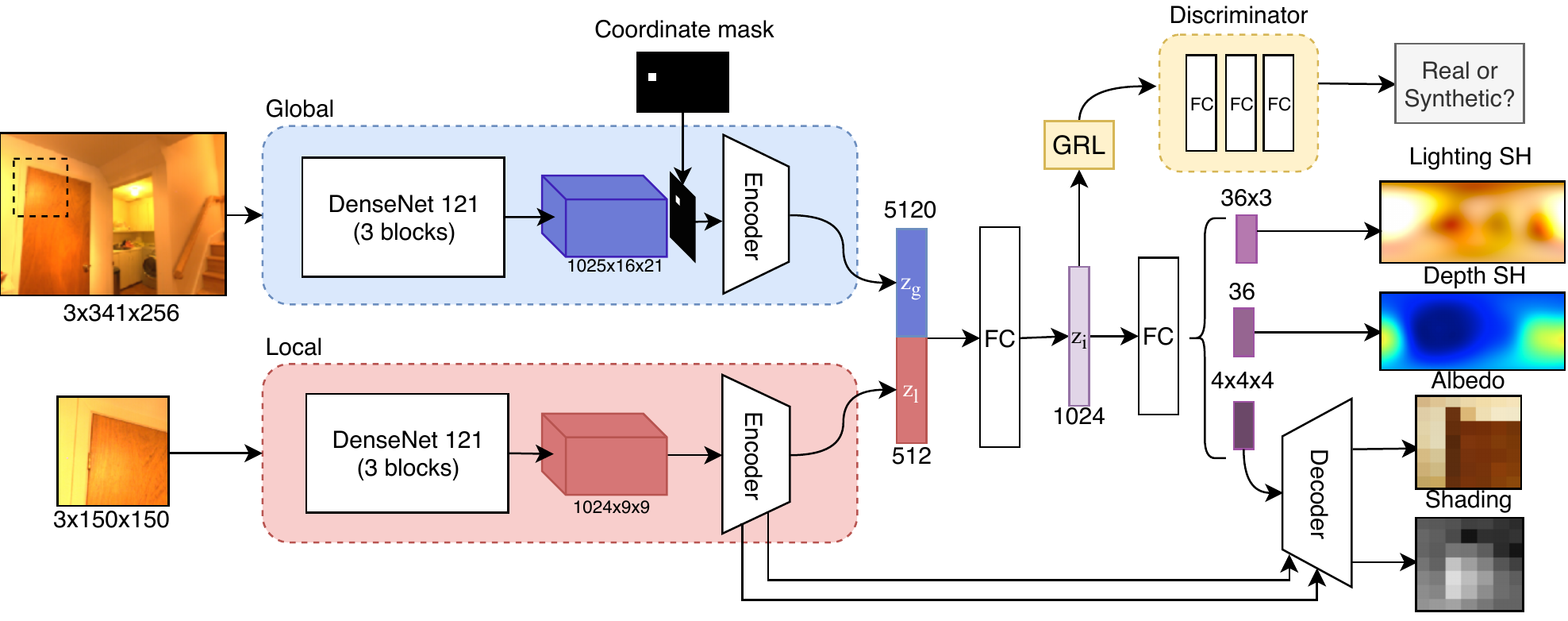}
\caption{The architecture of our neural network. In blue, the \emph{global path} where the full image is processed, and in red, the \emph{local path} where a patch of the image centered at the coordinate where we want to estimate the light probe is provided. In both paths, three pre-trained blocks of DenseNet~\cite{huang2017densely} and two Fire modules~\cite{iandola2016squeezenet} trained from scratch are used to obtain the local and global features. The features are combined with two fully connected layers to output the RGB spherical harmonics coefficients of the lighting and spherical harmonics coefficients of the depth. A decoder is jointly trained to regress the albedo and the shading of the local patch. We apply domain adaptation~\cite{ganin-icml-15} with a discriminator (yellow) and an adversarial loss on the latent vector to generalize to real images.}
\label{fig:network}
\end{figure*}

In order to learn to estimate local lighting, we need a large database of images and their corresponding illumination conditions (light probes) measured at several locations in the scene. Relying on panorama datasets such as~\cite{gardner-sigasia-17} unfortunately cannot be done since they do not capture local occlusions. While we provide a small dataset of real photographs for the evaluation of our approach (sec.~\ref{sec:dataset-real}), capturing enough such images to train a neural network would require a large amount of ressources. We therefore rely on realistic, synthetic data to train our neural network. In this section, we describe how we create our local light probe training data. 

\subsection{Rendering images}

As in \cite{zhang2016physically}, we use the SUNCG~\cite{song2016ssc} dataset for training. We do not use the Reinhard tonemapping algorithm~\cite{reinhard2002photographic} and instead use a simple gamma~\cite{li-cgintrinsics-18}. We now describe the corrections applied to the renders to improve their realism.

The intensity of the light sources in the SUNCG dataset are randomly scaled by a factor $e \sim \mathcal{U}[100, 500]$, where $\mathcal{U}[a,b]$ is a uniform distribution in the $[a,b]$ interval. Since many scenes have windows, we randomly sample a panorama from a dataset of 200 HDR outdoor panoramas~\cite{yannick-cvpr-19}. Each outdoor panorama is also randomly rotated about its vertical axis, to simulate different sun directions. We found the use of these panoramas to add significant realism to scenes with a window, providing realistic appearance and lighting distributions (see fig.~\ref{fig:dataset}). 

We render a total of 26,800 images, and use the same scenes and camera viewpoints as \cite{zhang2016physically}. Care is taken to split the training/validation dataset according to houses (each house containing many rooms). Each image is rendered at $640\!\times\!480$ resolution using the Metropolis Light Transport (MLT) algorithm of Mitsuba~\cite{Mitsuba}, with 512 samples.

\subsection{Rendering local light probes}

For each image, we randomly sample 4 locations in the scene to render the local light probes. The image is split into 4 quadrants, and a random 2D coordinate is sampled uniformly in each quadrant (excluding a 5\% border around the edges of the image). To determine the position of the virtual camera in order to render the light probe (the ``probe camera''), a ray is cast from the scene camera to the image plane, and the first intersection point with geometry is kept. From that point, we move the virtual camera 10cm away from the surface, along the normal, and render the light probe at this location. Note that the probe camera axes are aligned with those of the scene camera---only a translation is applied. 

Each light probe is rendered in the cubemap representation, rendering each of the 6 faces independently. While Mitsuba MLT can converge to a realistic image rapidly, it sometimes converges to a wrong solution that can negatively affect the ground truth probes. Thus, for each face we use the Bidirectional Path Tracing (BDPT) algorithm of Mitsuba~\cite{Mitsuba}, with 1,024 samples and render at $64\!\times\!64$ resolution. This takes, on average, 5 minutes to render all 6 faces of a light probe. In addition, we also render the depth at each probe. Fig.~\ref{fig:dataset} shows examples of images and their corresponding probes in our synthetic dataset. 

After rendering, scenes are filtered out to remove erroneous floor or wall area lights which are present in SUNCG. In addition, moving 10cm away from a surface may result in the camera being \emph{behind} another surface. To filter these probes out, we simply threshold based on the mean light probe intensity (a value of 0.01 was found empirically). Finally, we fit 5th order SH coefficients to the rendered light and depth probes to obtain ground truth values to learn. 
\section{Learning to estimate local indoor lighting}
\label{sec:learning}

\subsection{Main architecture for lighting estimation}
\label{sec:architecture}

We now describe our deep network architecture to learn spatially-varying lighting from an image. Previous work has shown that global context can help in conducting reasoning on lighting~\cite{gardner-sigasia-17}, so it is likely that a combination of global and local information will be needed here. 

Our proposed architecture is illustrated in fig.~\ref{fig:network}. We require an input RGB image of $341\times256$ resolution and a specific coordinate in the image where the lighting is to be estimated. The image is provided to a ``global'' path in the CNN. A local patch of $150\times150$ resolution, centered on that location, is extracted and fed to a ``local'' path. 

The global path processes the input image via the three first blocks of a pretrained DenseNet-121 network to generate a feature map. A binary coordinate mask, of spatial resolution $16\times21$, with the elements corresponding to the local patch set to 1 and 0 elsewhere, is concatenated as an additional channel to the feature map. The result is fed to an encoder, which produces a 5120-dimensional vector $\mathbf{z}_g$. The local path has a similar structure. It processes the input patch with a pretrained DenseNet-121 network to generate a feature map, which is fed to an encoder and produces a 512-dimensional vector $\mathbf{z}_l$.
Both global and local encoders share similar structures and use Fire modules~\cite{iandola2016squeezenet}. With the fire-$x$-$y$ notation, meaning that the module reduces the number of channels to $x$ before expanding to $y$, the encoders have the following structure: fire-512-1024/fire-256-256 for the global stream, and fire-512-512/fire-128-128 for the local stream. The Fire modules have ELU activations and are followed by batch normalization. 

The vectors $\mathbf{z}_g$ and $\mathbf{z}_l$ coming from the global and local paths respectively are concatenated and processed by a fully-connected (FC) layer of 1024 neurons to produce the latent vector $\mathbf{z}_i$. The 5th-order SH coefficients in RGB are then predicted by another FC layer of dimensionality $36\times3$. We use an MSE loss on the SH coefficients:
\begin{equation} 
\mathcal{L}_\text{i-sh} = \frac{1}{36\times3} \sum_{c=1}^3 \sum_{l=0}^{4} \sum_{m=-l}^l (i_{l,c}^{m*} - i_{l,c}^m)^2 \,,
\label{eqn:loss-ish}
\end{equation}
where $i_{l,c}^m$ ($i_{l,c}^{m*}$) are the predicted (w.r.t. ground truth) SH coefficients for the $c$-th color channel (in RGB).

\subsection{Learning additional subtasks}
\label{sec:learning-subtasks}

It has recently been shown that similar tasks can benefit from joint training~\cite{zamir2018taskonomy}. We now describe additional branches and losses that are added to the network to learn these related tasks, and in sec.~\ref{sec:validation-synthetic} we present an ablation study to evaluate the impact of each of these subtasks. 

\paragraph{Learning low-frequency probe depth} Since lighting is affected by local visibility---for example, lighting under a table is darker because the table occludes overhead light sources---we ask the network to also predict SH coefficients for the low-frequency probe depth (fig.~\ref{fig:dataset}). To do so, we add another 36-dimensional output to the last FC layers after the $\mathbf{z}_i$ vector (fig.~\ref{fig:network}, right). The loss for this branch is the MSE on the depth SH coefficients:
\begin{equation} 
\mathcal{L}_\text{d-sh} = \frac{1}{36} \sum_{l=0}^{4} \sum_{m=-l}^l (d_l^{m*} - d_l^m)^2 \,,
\label{eqn:loss-depth}
\end{equation}
where $d_l^m$ ($d_l^{m*}$) are the spherical harmonics coefficients for the probe depth (w.r.t. ground truth).

\paragraph{Learning patch albedo and shading} To help disambiguate between reflectance and illumination, we also ask the network to decompose the local patch into its reflectance and shading intrinsic components. For this, we add a 3-layer decoder that takes in a $4\times4\times4$ vector from the last FC layer in the main branch, and reconstructs $7\times7$ pixel resolution (color) albedo and (grayscale) shading images. We do this at low-resolution because we want to summarily disambiguate color and intensity between local reflectance and illumination. The encoder is composed of the following 3 layers: conv3-128, conv3-128 and conv1-4 (where conv$x$-$y$ denotes a convolution layer of $y$ filters of dimension $x \times x$) respectively. The first two layers have ELU activations followed by batch norm. A $2\times$ upsample is applied after the first convolution to allow skip links~\cite{ronneberger2015u} between the local encoder and decoder. The last layer has a sigmoid activation function for the 3 albedo channels. We define the losses on reflectance and shading as: 
\begin{equation}
\begin{split}
\mathcal{L}_\text{rs-mse} &= \frac{1}{N} \sum_{i=1}^N (\mathbf{R}^*_i - \mathbf{R}_i)^2 + (\mathbf{S}^*_i - \mathbf{S}_i)^2 \\
\mathcal{L}_\text{rs-recons} &= \frac{1}{N} \sum_{i=1}^N \left(\mathbf{P}^*_i - (\mathbf{R}_i + \mathbf{S}_i) \right)^2
\end{split} \,,
\label{eqn:loss-rs}
\end{equation}
where $\mathbf{R}_i$ ($\mathbf{R}_i^*$) and $\mathbf{S}_i$ ($\mathbf{S}_i^*$) denote the log-reflectance prediction (resp. ground truth) and log-shading prediction (resp. ground truth) respectively, and $\mathbf{P}^*_i$ is the input patch.

\paragraph{Adapting to real data} We apply unsupervised domain adaptation~\cite{ganin-icml-15} to adapt the model trained on synthetic SUNCG images to real photographs. This is done via a discriminator connected to the $\mathbf{z}_i$ latent vector by a gradient reversal layer (GRL). The discriminator, illustrated in the top-right of fig.~\ref{fig:network}, is composed of 3 FC layers of (64, 32, 2) neurons respectively, with the ELU activation and the first two layers followed by batch normalization. We train the discriminator with the cross-entropy loss: 
\begin{equation}
\mathcal{L}_\text{da} = -\sum_{i=1}^N r^*_i \log r_i \,,
\end{equation}
where $r_i$ ($r^*_i$) is the discriminator output (resp. ground truth). 

\subsection{Training}

The overall loss for our neural network is a linear combination of the individual losses introduced in sec.~\ref{sec:learning-subtasks}: 
\begin{equation}
\mathcal{L} = \mathcal{L}_\text{i-sh} + \mathcal{L}_\text{d-sh} + \mathcal{L}_\text{rs-mse} + \mathcal{L}_\text{rs-recons} + \lambda \mathcal{L}_\text{da}\,.
\label{eqn:loss}
\end{equation}
Here, $\lambda = 2 / (1 + \exp(-8 n/60)) - 1$, where $n$ is the epoch, is a weight controlling the importance of the domain adaptation loss, which becomes asymptotically closer to 1 as the number of training epochs $e$ increases. We train the network from scratch (apart for the DenseNet-121 blocks which are pretrained on ImageNet) using the Adam optimizer with $\beta=(0.9,0.999)$. We employ a learning rate of $10^{-4}$ for the first 60 epochs, and of $10^{-5}$ for 30 additional epochs. 

The network is trained on a combination of synthetic and real data. Each minibatch, of total size 20, is composed of 50\% synthetic, and 50\% real data. For the synthetic data, all losses in eq.~(\ref{eqn:loss}) are activated. In total, we use 83,812 probes from 24,000 synthetic images for training, and 9,967 probes from 2,800 images for validation (to monitor for over-fitting). We augment the synthetic data (images and probes) at runtime, while training the network, using the following three strategies: 1) horizontal flip; 2) random exposure factor $f \sim \mathcal{U}[0.16, 3]$, where $\mathcal{U}[a,b]$ is the uniform distribution in the $[a,b]$ interval as in sec.~\ref{sec:dataset}; and 3) random camera response function $f(x) = x^{1/\gamma}, \gamma \sim \mathcal{U}[1.8, 2.4]$. The real data is composed of crops extracted from the Laval Indoor HDR dataset~\cite{gardner-sigasia-17}. Since we do not have any ground truth for the real data, we only employ the $\mathcal{L}_\text{da}$ loss in this case. 

\begin{table}[!t]
    \footnotesize
    \centering
    \begin{tabular}{ccccc}
     SH Degree	  	& \makecell{Global \\ (w/o mask)} 	& \makecell{Global \\(w mask)} 		&  Local	& \makecell{Local + Global \\(w mask)} 	\\
    \cmidrule{2-5}
     0 				& 0.698								& 0.563 							& 0.553								& \textbf{0.520}							\\
     1				& 0.451								& 0.384 							& 0.412								& \textbf{0.379}							\\
     2--5			& 0.182								& \textbf{0.158} 					& 0.165								& 0.159							\\
    
    \end{tabular}
    \caption{Ablation study on the network inputs. The mean absolute
    error (MAE) of each SH degree on the synthetic test set are reported. ``Global w/o mask'' takes the full image without any probe position information. Two types of local information are evaluated: ``Global (w mask)'' receives the full image and the coordinate mask of the probe position, and ``Local'' receives a patch around the probe position. Our experiments show that using both local information and the full image (``Local + Global (w mask'') reduces the error.}
    \label{tab:ablation_input}
\end{table}

\begin{table}[!t]
    \footnotesize
    \centering
    \begin{tabular}{ccccc}
    \centering
     SH Degree	  	&  
     $\mathcal{L}_\text{i-sh}$ & 
     $+ \mathcal{L}_\text{d-sh}$ &   
     \makecell{$+ \mathcal{L}_\text{rs-mse} $ \\ $+\mathcal{L}_\text{rs-recons}$}	& 
     All \\
    \cmidrule{2-5}
     0 				& 0.520								& 0.511 							& 0.472								& \textbf{0.449}					\\
     1				& 0.379								& 0.341 							& 0.372								& \textbf{0.336}					\\
     2--5			& 0.159								& 0.149 							& 0.166								& \textbf{0.146}					\\
    \cmidrule{2-5}
    Degree 1 angle & 0.604 &  0.582 & 0.641 & \textbf{0.541} \\
    \end{tabular}
    
    \caption{Comparing the mean absolute error (MAE) of the lighting SH degrees for each loss from 10,000 synthetic test probes. Estimating the low frequency depth at the probe position improves the directional degrees of the SH while providing minimal gain on the ambient light (degree 0). The albedo and shading losses improve the ambient light estimation. ``Degree 1 angle'' represents the angular error of the first order SH. Training the network on all of these tasks achieves better results for all of the degrees.}
    \label{tab:ablation_loss}
  \end{table}
\section{Experimental validation}

We now present an extensive evaluation of our network design as well as qualitative and quantitative results on a new benchmark test set. We evaluate our system's accuracy at estimating 5th order SH coefficients. We chose order 5 after experimenting with orders ranging from 3 to 8, and empirically confirming that order 5 SH lighting gave us a practical trade-off between rendering time and visual quality (including shading and shadow softness). In principle, our network can be easily extended to infer higher order coefficients.  

\subsection{Validation on synthetic data}
\label{sec:validation-synthetic}

A non-overlapping test set of 9,900 probes from 2,800 synthetic images (sec.~\ref{sec:dataset}) is used to perform two ablation studies to validate the design choices in the network architecture (sec.~\ref{sec:architecture}) and additional subtasks (sec.~\ref{sec:learning-subtasks}). 

First, we evaluate the impact of having both global and local paths in the network, and report the mean absolute error (MAE) in SH coefficient estimation in tab.~\ref{tab:ablation_input}. For this experiment, the baseline (``Global (w/o mask)'') is a network that receives only the full image, similar to Gardner et al.~\cite{gardner-sigasia-17}. Without local information, the network predicts the average light condition of the scene and fails to predict local changes, thus resulting in low accuracy. Lower error is obtained by concatenating the coordinate mask to the global DenseNet feature map (``Global (w mask)''). Interestingly, using only the local path (red in fig.~\ref{fig:network}, ``Local'' in tab.~\ref{tab:ablation_input}) gives better accuracy than the global image, hinting that local lighting can indeed be quite different from the global lighting. Using both types of local information, i.e. the coordinate mask in the global path and the local patch, lowers the error further (``Local + Global (w mask)'' in tab.~\ref{tab:ablation_input}).

\begin{figure}
  \centering
  \setlength{\tabcolsep}{1pt}
  \begin{tabular}{cc}
  \includegraphics[width=0.5\linewidth]{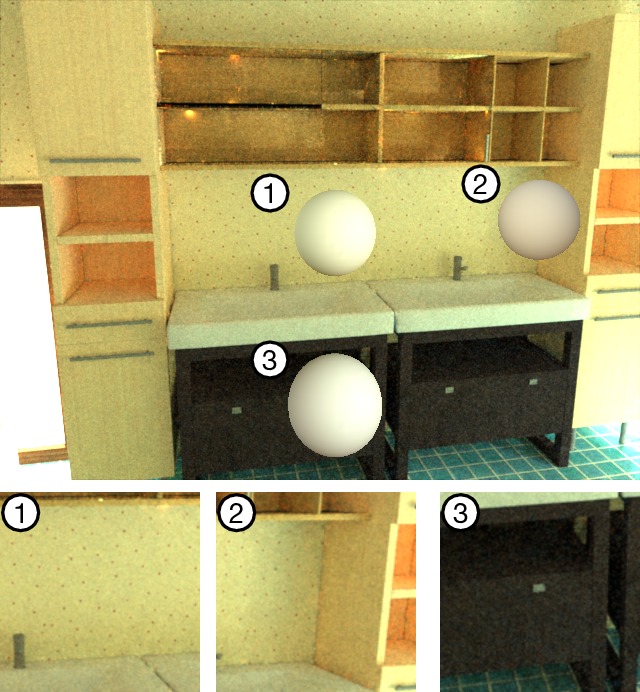} &
  \includegraphics[width=0.5\linewidth]{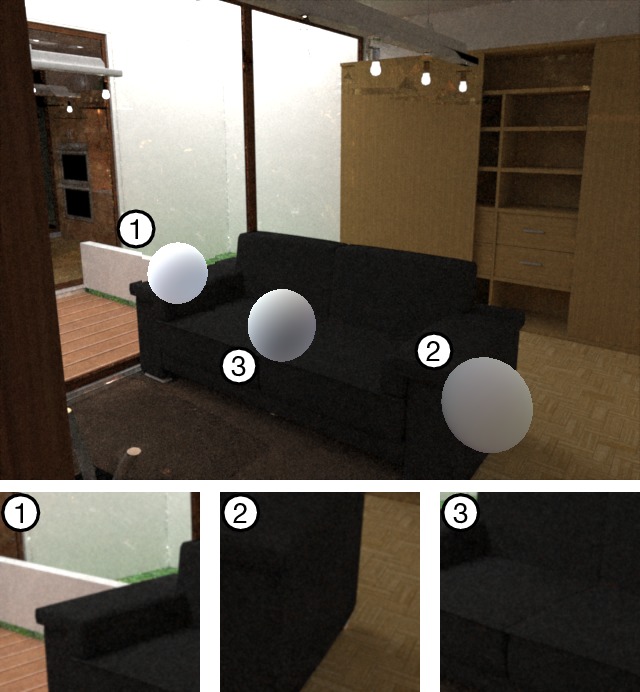}
  \end{tabular}
  \caption{Qualitative examples of robustness to albedo changes. Our network adapts to the changes in albedo in the scene, and does not strictly rely on average patch brightness to estimate ambient lighting. We demonstrate three estimations: (1) a reference patch, (2) a patch that has similar brightness as the reference, but different lighting; and (3) a patch that has similar lighting as the reference, but different brightness. Notice how our network adapts to these changes and predicts coherent lighting estimates.
  }
  \label{fig:albedo}
  \end{figure}

Second, tab.~\ref{tab:ablation_loss} shows that learning subtasks improves the performance for the light estimation task~\cite{zamir2018taskonomy}. In these experiments, the entire network with only the loss on SH coefficients $\mathcal{L}_\text{i-sh}$ from eq.~(\ref{eqn:loss-ish}) is taken as baseline. Activating the MSE loss on the low frequency probe depth $\mathcal{L}_\text{d-sh}$ from eq.~(\ref{eqn:loss-depth}) significantly improves the directional components of the SH coefficients, but has little impact on the degree 0. Conversely, training with an albedo/shading decomposition task (loss functions $\mathcal{L}_\text{rs-mse}$ and $\mathcal{L}_\text{rs-recons}$ from eq.~(\ref{eqn:loss-rs})) improves the ambient light estimation (SH degree 0), but leaves the directional components mostly unchanged. In fig.~\ref{fig:albedo},we show that our network is able to differentiate local reflectance from shading changes and does not only rely on mean patch color. Combining all subtasks improves both the ambient and the directional predictions of the network.

\subsection{A dataset of real images and local light probes}
\label{sec:dataset-real}

\begin{figure*}[!t]
  \centering
  \setlength{\tabcolsep}{6pt}
  \begin{tabular}{c|c}
  \includegraphics[width=.48\linewidth]{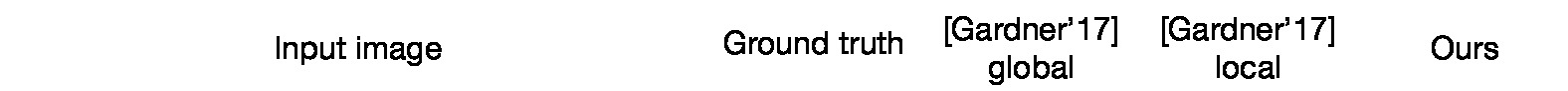} & 
  \includegraphics[width=.48\linewidth]{figures/eval-qual/eval-qual-labels.jpg} \\
  \includegraphics[width=.48\linewidth]{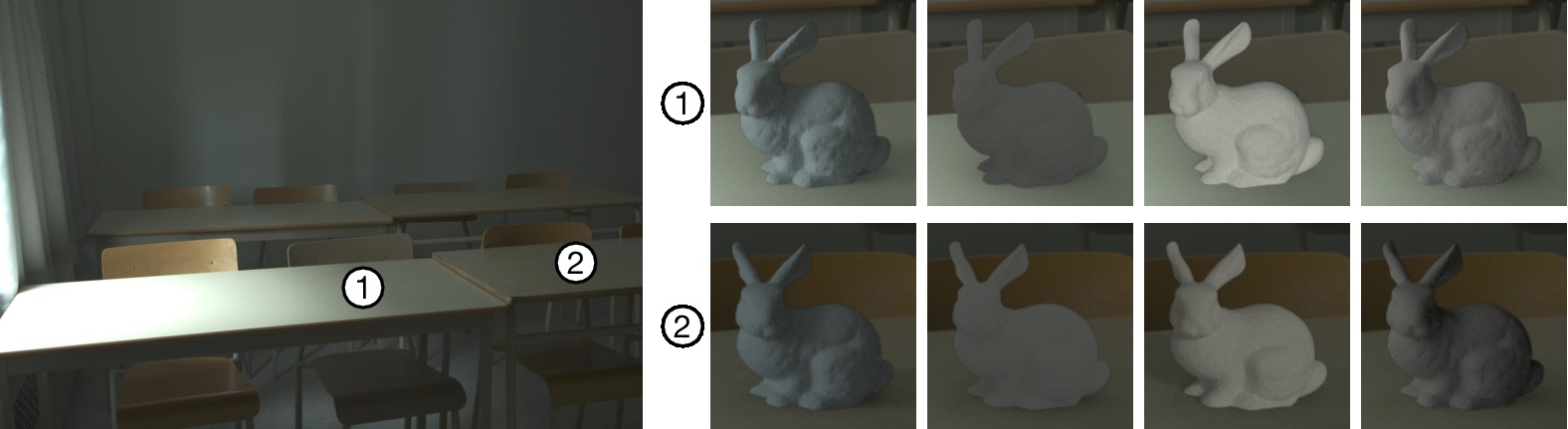} & 
  \includegraphics[width=.48\linewidth]{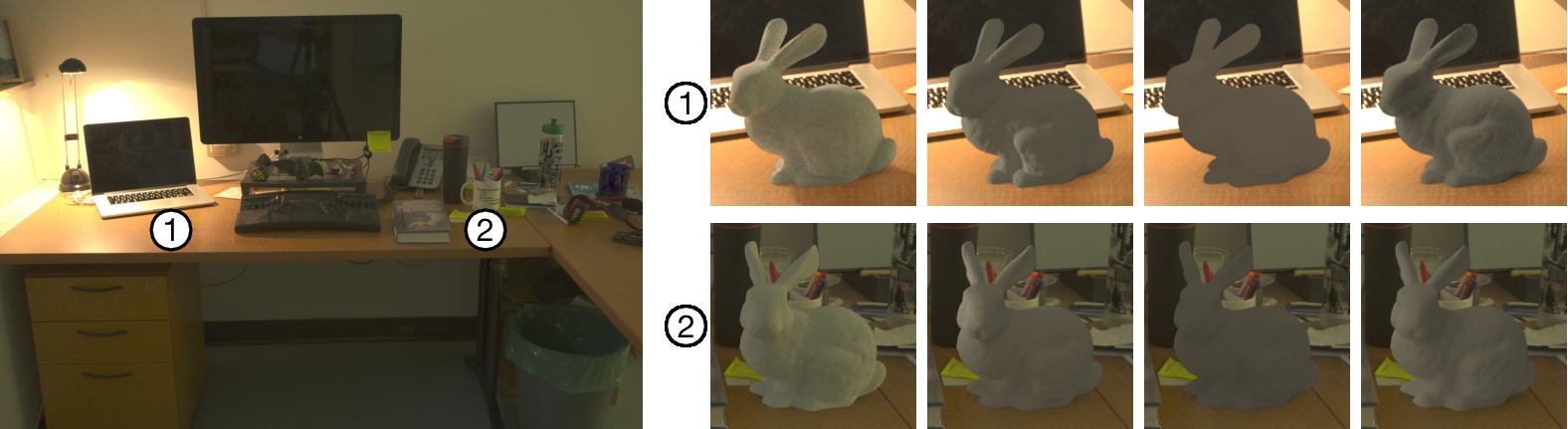} \\
  \includegraphics[width=.48\linewidth]{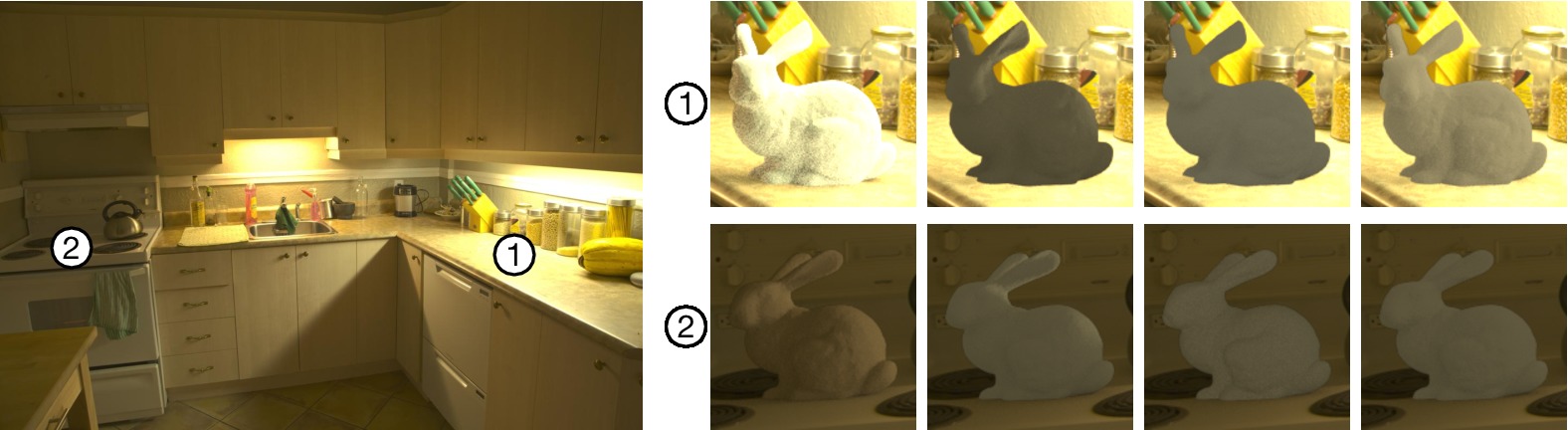} & 
  \includegraphics[width=.48\linewidth]{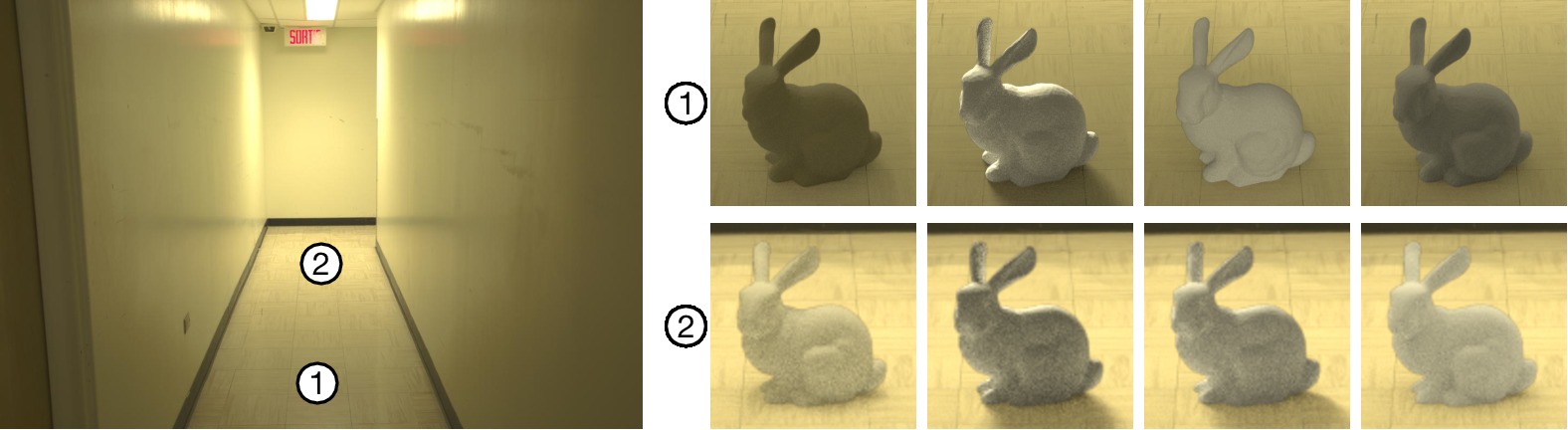} \\
  \includegraphics[width=.48\linewidth]{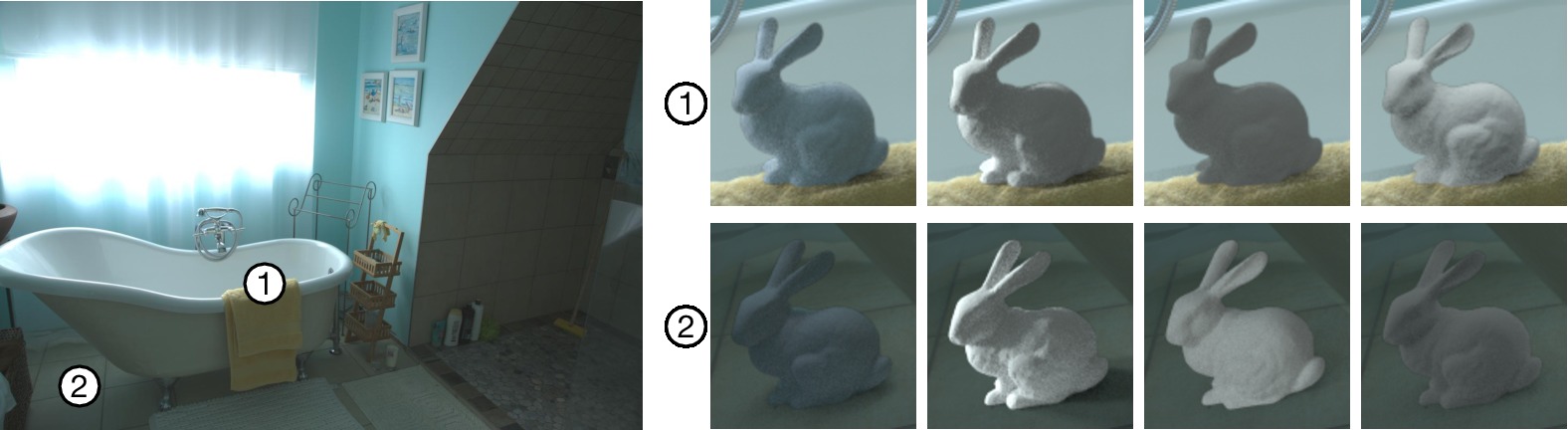} & 
  \includegraphics[width=.48\linewidth]{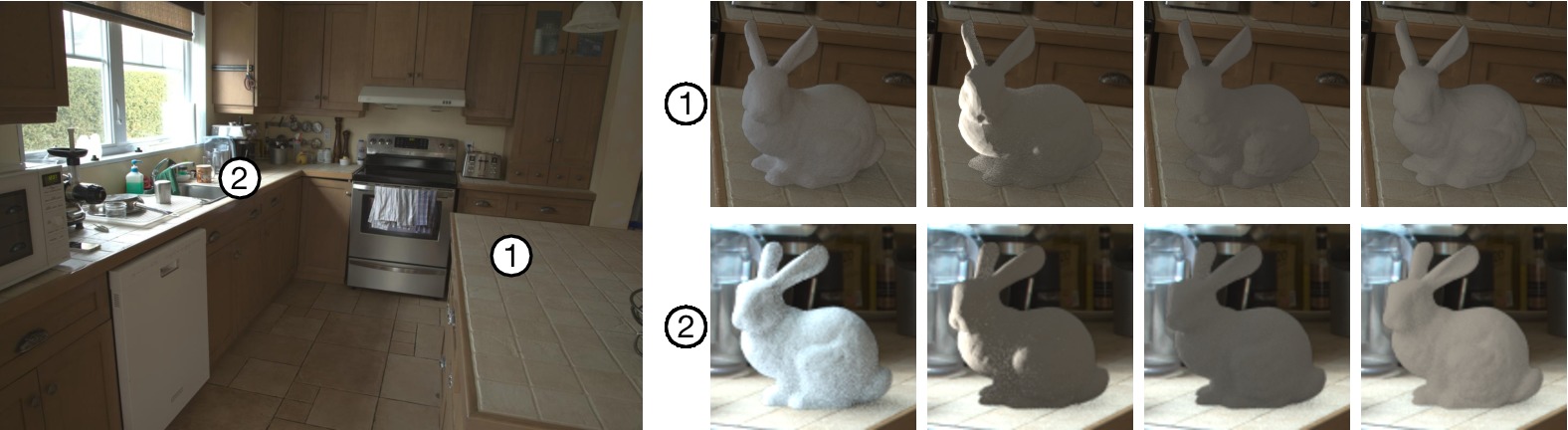} \\
  \end{tabular}
  \caption{Qualitative comparative results on our dataset of real images and local light probes. These results were also part of our user study.}
  \label{fig:results-qual}
\end{figure*}

To validate our approach, we captured a novel dataset of real indoor scenes and corresponding, spatially-varying light probes (see fig.~\ref{fig:results-qual}). The images were captured with a Canon EOS 5D mark III and a 24--105 mm lens on a tripod. The scenes are first captured in high dynamic range by merging 7 bracketed exposures (from 1/8000s to 8s) with an f/11 aperture. For each scene, an average of 4 HDR light probes are subsequently captured by placing a 3-inch diameter chrome ball~\cite{debevec2008recovering} at different locations, and shooting the entire scene with the ball in HDR once more. The metallic spheres are then segmented out manually, and the corresponding environment map rotated according to its view vector with respect to the center of projection of the camera. In all, a total of 20 indoor scenes and 79 HDR light probes were shot. In the following, we use the dataset to compare our methods quantitatively, and through a perceptual study.

\begin{table}[!t]
  \centering
  \footnotesize\textbf{}
  \begin{tabular}{ccc}
    & w/o domain adaptation		& domain adaptation 	\\
  \cmidrule(ll){2-3}
  RMSE 		& {0.051 $\pm$ 0.011}	&	\textbf{0.049} $\pm$ \textbf{0.006} \\
  siRMSE 		& {0.072 $\pm$ 0.009}	&	\textbf{0.062} $\pm$ \textbf{0.005}  \\
  \end{tabular}
  \vspace{.5em}
  \caption{Comparing the effect of domain adaptation loss on all the probes from the real images using a relighting error. Domain adaptation slightly improves the performances of the method.}
  \label{tab:da}
\end{table}
  
  \begin{table}[!t]
  \centering
  \footnotesize
  \begin{tabular}{clccc}
   & & All 		& Center & Off-center  	\\
  \cmidrule(ll){3-5}
  \parbox[t]{2mm}{\multirow{3}{*}{\rotatebox[origin=c]{90}{RMSE}}} 
  & global-\cite{gardner-sigasia-17} 		& {0.081 $\pm$ 0.015}     &   {0.079 $\pm$ 0.021}  &  {0.086 $\pm$ 0.019}  \\
  & local-\cite{gardner-sigasia-17} 		& {0.072 $\pm$ 0.013}     &   {0.086 $\pm$ 0.027}   &  {0.068 $\pm$ 0.019} 				\\
  &Ours		& \textbf{0.049} $\pm$ \textbf{0.006}  &  \textbf{0.049} $\pm$ \textbf{0.019}  &  \textbf{0.051} $\pm$ \textbf{0.012} 	\\
  \cmidrule(ll){3-5}
  \parbox[t]{2mm}{\multirow{3}{*}{\rotatebox[origin=c]{90}{siRMSE}}} 
  & global-\cite{gardner-sigasia-17} 		& {0.120 $\pm$ 0.013}     &   {0.124 $\pm$ 0.018}   &  {0.120 $\pm$ 0.031}  \\
  & local-\cite{gardner-sigasia-17} 						& {0.092 $\pm$ 0.017}     &   {0.120 $\pm$ 0.035}   &  {0.084 $\pm$ 0.016}  \\
  &Ours					& \textbf{0.062} $\pm$ \textbf{0.005}     &   \textbf{0.072} $\pm$ \textbf{0.011}   &  \textbf{0.055} $\pm$ \textbf{0.009} \\
  \end{tabular}
  
  \vspace{.5em}
  \caption{Comparing the relighting error between each method. We show results for all probes, \emph{center} probes that are close to the center of the image and not affected by the local geometry (e.g. shadows) or close to a light source, and \emph{off-center} probes. We report the RMSE and si-RMSE~\cite{grosse-iccv-09} and their 95\% confidence intervals (computed using bootstrapping). As expected, global-\cite{gardner-sigasia-17} has a lower RMSE error for the center probes. Our method has a lower error on both metrics and is constant across all type of probes.}
  \label{tab:result}
\end{table}

\subsection{Quantitative comparison on real photographs}
\label{sec:quant_comp}

We use the real dataset from sec.~\ref{sec:dataset-real} to validate the domain adaptation proposed in sec.~\ref{sec:learning-subtasks} and to compare our method against two versions of the approach of Gardner et al.~\cite{gardner-sigasia-17}, named \emph{global} and \emph{local}. The global version is their original algorithm, which receives the full image as input and outputs a single, global lighting estimate. For a perhaps fairer comparison, we make their approach more local by giving it as input a crop containing a third of the image with the probe position as close as possible to the center. 

\paragraph{Relighting error} We compare all methods by rendering a diffuse bunny model with the ground truth environment map, with the algorithms outputs, and compute error metrics on the renders with respect to ground truth. Note that, since our method outputs SH coefficients, we first convert them to an environment map representation, and perform the same rendering and compositing technique as the other methods. Table~\ref{tab:da} demonstrates that training with the domain adaptation loss slightly improves the scores on the real dataset. A comparison against~\cite{gardner-sigasia-17} is provided in tab.~\ref{tab:result}. To provide more insight, we further split the light probes in the dataset into two different categories: the \emph{center} and \emph{off-center} probes. The center probes were determined, by manual inspection, to be those close to the center of the image, and not affected by the local geometry or close light sources. All center and off-center probes are presented in the supplementary material. Since it is trained to estimate the lighting approximately in the middle of the image, the RMSE of global-\cite{gardner-sigasia-17} is slightly lower on the center probes than the off-center ones. We note that our method outperforms both versions of \cite{gardner-sigasia-17}. 

\paragraph{User study}
We further conduct a user study to evaluate whether the quantitative results obtained in the previous section are corroborated perceptually. For each of the 3 techniques (ours, global-\cite{gardner-sigasia-17}, local-\cite{gardner-sigasia-17}), we show the users pairs of images: the reference image rendered with the ground truth light probe, and the result rendered with one of the lighting estimates. Each user is presented with all of the 20 scenes, and for each scene a random probe and a random technique is selected. Example images used are shown in fig.~\ref{fig:results-qual}. 

The study was conducted using Amazon Mechanical Turk. The users were permitted to do the test at most twice to prevent statistical bias, and 2 sentinels (obvious choices) were inserted to filter out bad responses. A total of 144 unique participants took part in the study, resulting in an average of 20 votes per light probe. Tab.~\ref{tab:user_study} shows the results of our user study, including results split by center and off-center probes (as defined in sec.~\ref{sec:quant_comp}). In all, our method achieves a confusion of 35.8\% (over a maximum of 50\%), compared to 28\% and 31\% for the local and global versions of \cite{gardner-sigasia-17}. We note that global-\cite{gardner-sigasia-17} outperforms ours slightly on the ``center'' probes with a 39.8\% vs 38.3\% confusion. We attribute this performance to the fact that this technique is trained specifically for this kind of scenario, and its non-parametric output has the tendency to predict sharper shadows than ours---something that seems to be appreciated by users. On the other hand, the performance of global-\cite{gardner-sigasia-17} severely decreases on ``off-center'' probes with 27.1\% confusion. In contrast, our method still maintains a high confusion with 34.5\%. The local-\cite{gardner-sigasia-17} does not seem to work well in either case, probably due to the lack of context. 

\begin{table}[!t]
  \centering
  \begin{tabular}{lccc}
    & All 						& Center 	&  Off-center 	\\
  \cmidrule{2-4}
  global-\cite{gardner-sigasia-17} 		&  31.0\% 	        &	\textbf{39.8}\%          &  27.1\%\\
  local-\cite{gardner-sigasia-17} 		& 	28.0\%	        &	25.2\%                  &  29.5\% \\
  Ours 	                                & \textbf{35.8}\%	&	38.3\%                  &  \textbf{34.5}\% \\
  
  \end{tabular}
  \caption{Results of a user study on our dataset of real images and local light probes. Users were asked to compare between objects relit with reference ground truth lighting and a randomly chosen method. Overall, users had a higher confusion with our method (35.8\%) than the state-of-the-art (at best, 31.0\%). global-\cite{gardner-sigasia-17} has a high confusion rate on the center probes where it was specifically trained. Note that perfect performance is 50\%.
  }
  \label{tab:user_study}
  \end{table}

\paragraph{Comparison to Barron and Malik~\cite{barron2013rgbd}}

The method of Barron and Malik~\cite{barron2013rgbd} takes as input, a single RGB-D image, and returns spatially-varying 2nd order SH lighting for the scene. The results of the two algorithms on images from the NYU-v2 dataset~\cite{Silberman:ECCV12} are provided in fig.~\ref{fig:barron} and in the supplementary material. We note that their method does not capture abrupt changes in ambient lighting caused by local geometry which our method handles well. In addition, their method requires depth information and takes up to 1.5 hour to run for a single image; our method runs on RGB images and takes less than 20ms to process an image.

\begin{figure}
  \centering
  \footnotesize
  \setlength{\tabcolsep}{1pt}
  \begin{tabular}{cc}
  \includegraphics[trim={30px 30px 30px 30px},clip,width=.5\linewidth]{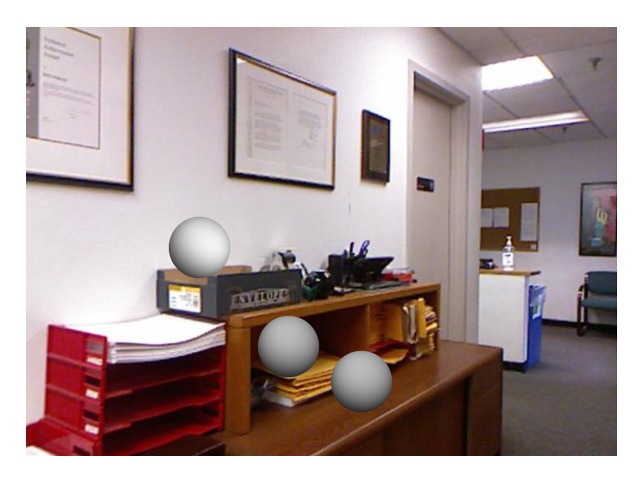} & 
  \includegraphics[trim={30px 30px 30px 30px},clip,width=.5\linewidth]{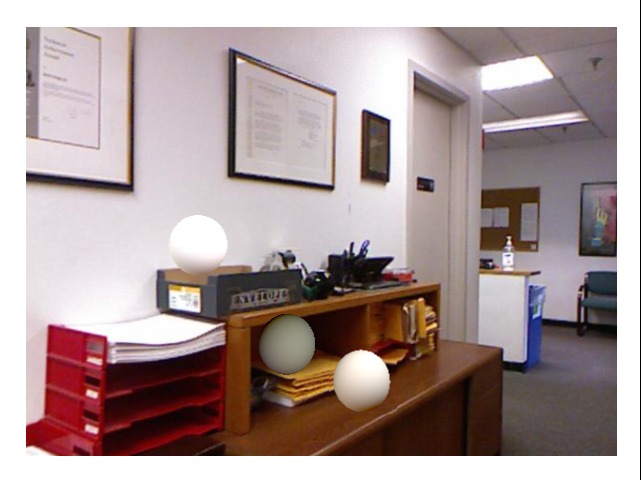} \\
  (a) Barron and Malik~\cite{barron2013rgbd} & (b) Ours \\
  \end{tabular}
  \vspace{.25em}
  \caption{Qualitative comparison to Barron and Malik~\cite{barron2013rgbd} on the NYU-v2 dataset~\cite{Silberman:ECCV12}. While their approach yields spatially-varying SH lighting, it typically produces conservative estimates that do not capture the spatial variation in lighting accurately. In contrast, our method requires only  RGB input, runs in real-time, and yields more realistic lighting estimates.
  }
  \label{fig:barron}
\end{figure}

\paragraph{Failure cases}

\begin{figure}
\centering
\begin{tabular}{cc}

\includegraphics[width=0.45\linewidth]{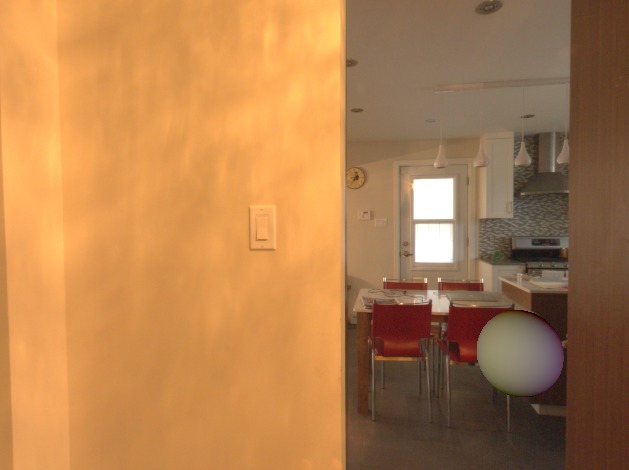} &
\includegraphics[width=0.45\linewidth]{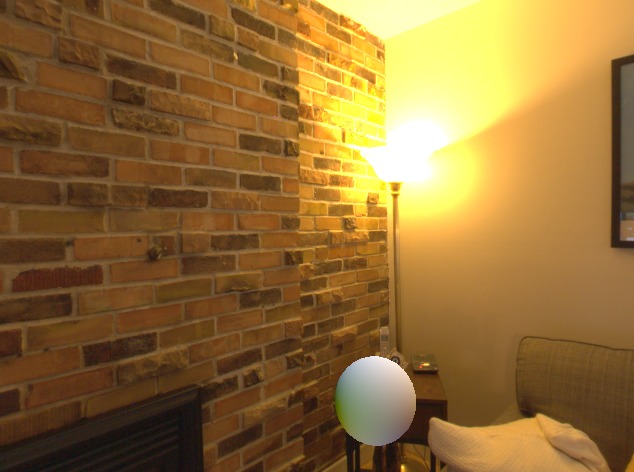} \\*[.5em]
\end{tabular}

\caption{On some scenes our algorithm predicts lighting with a wrong hue (typically purple or green).}
\label{fig:fails}
\end{figure}

While our method produces accurate lighting under a wide array of conditions, in some cases, we observe incorrect hue shifts (see fig.\ref{fig:fails}). We hypothesize that this could be a result of the differences between the light and camera response distributions of our synthetic training and real test sets; we belive it can be remedied through additional regularization or dataset enrichment.

\section{Real-time lighting estimation for AR}

\begin{figure}
\centering
\footnotesize
\setlength{\tabcolsep}{1pt}
\begin{tabular}{cc}
\includegraphics[width=.5\linewidth]{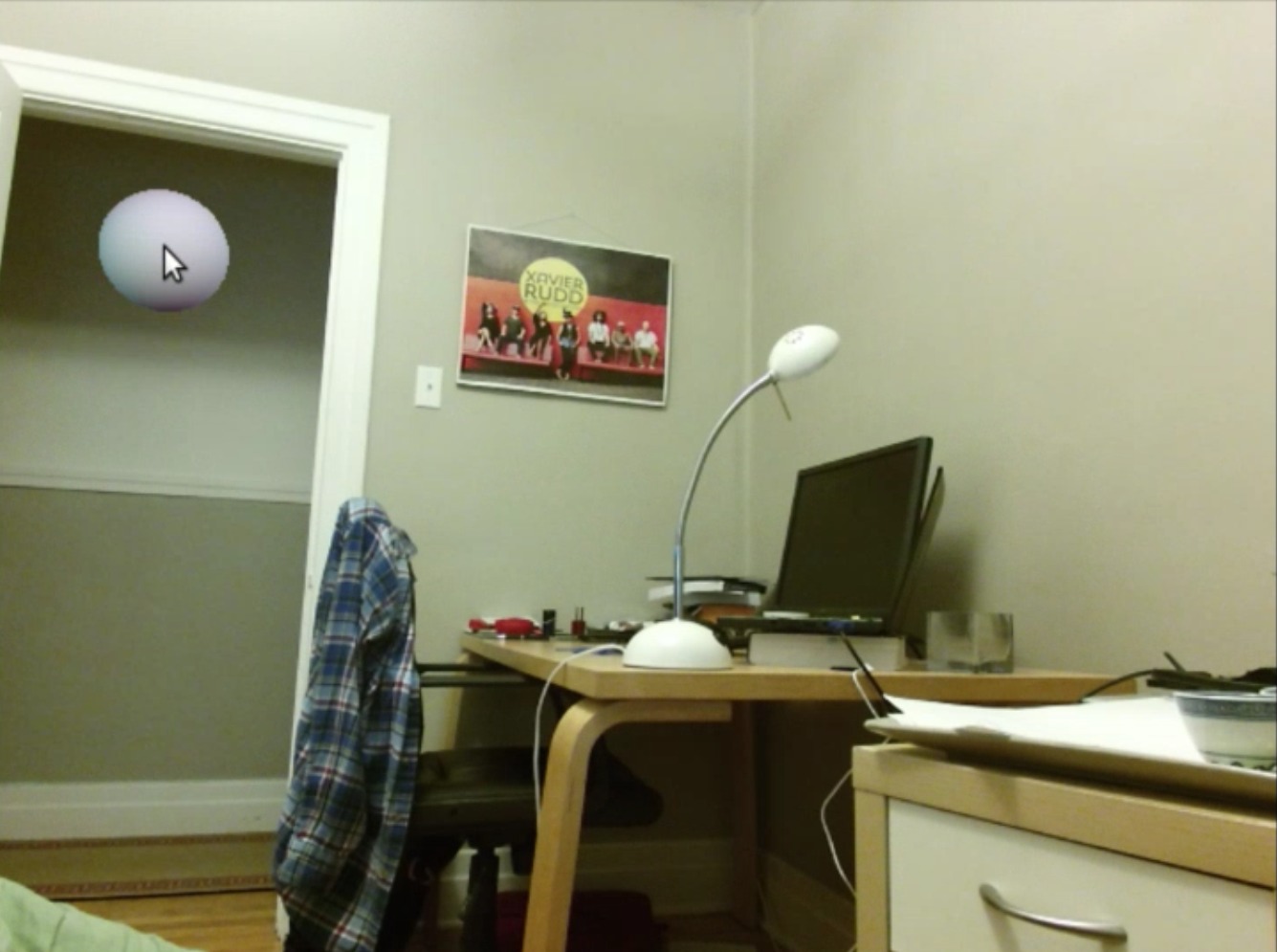} & 
\includegraphics[width=.5\linewidth]{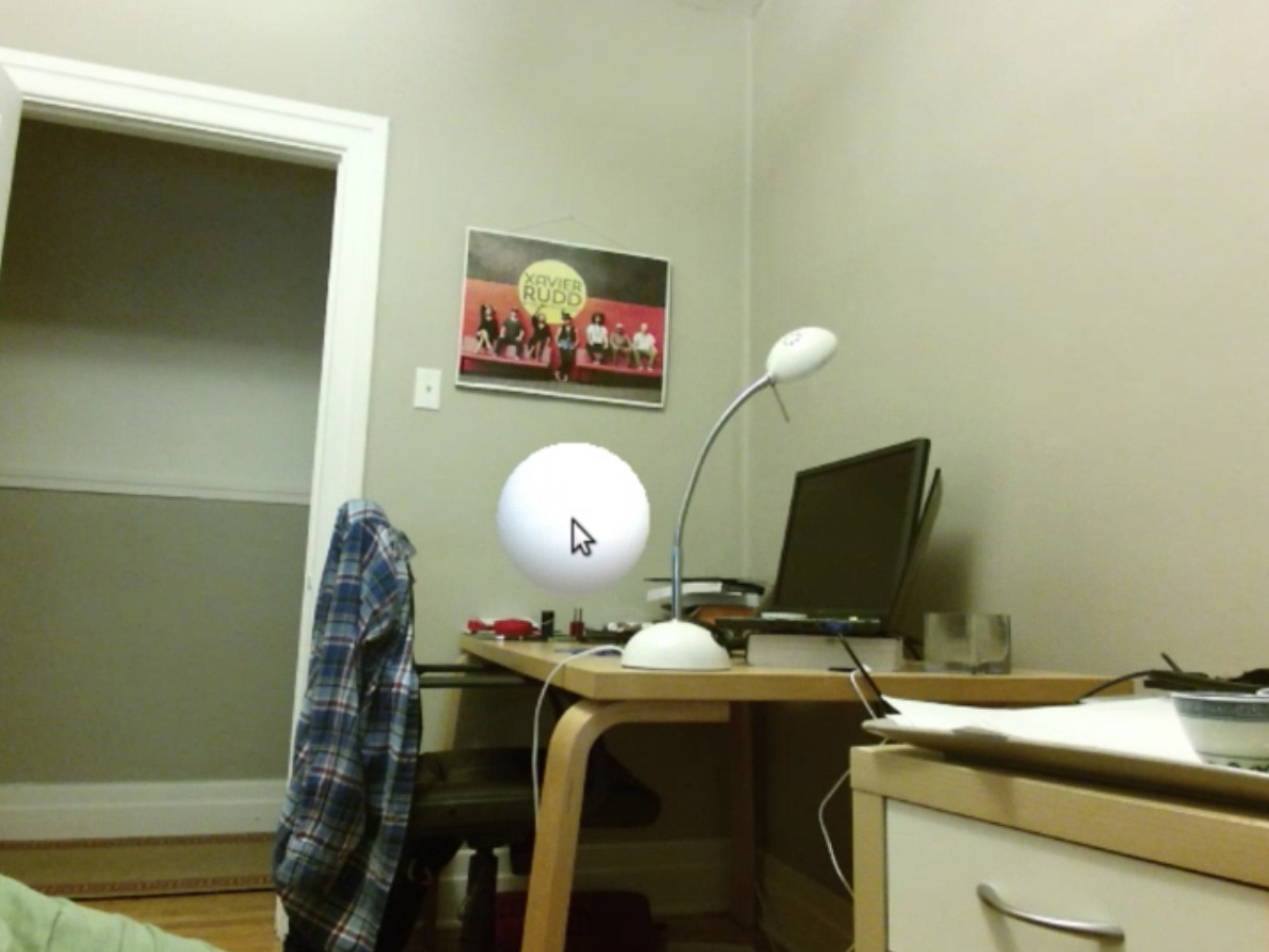} \\
\multicolumn{2}{c}{(a) Moving the object} \\
\includegraphics[width=.5\linewidth]{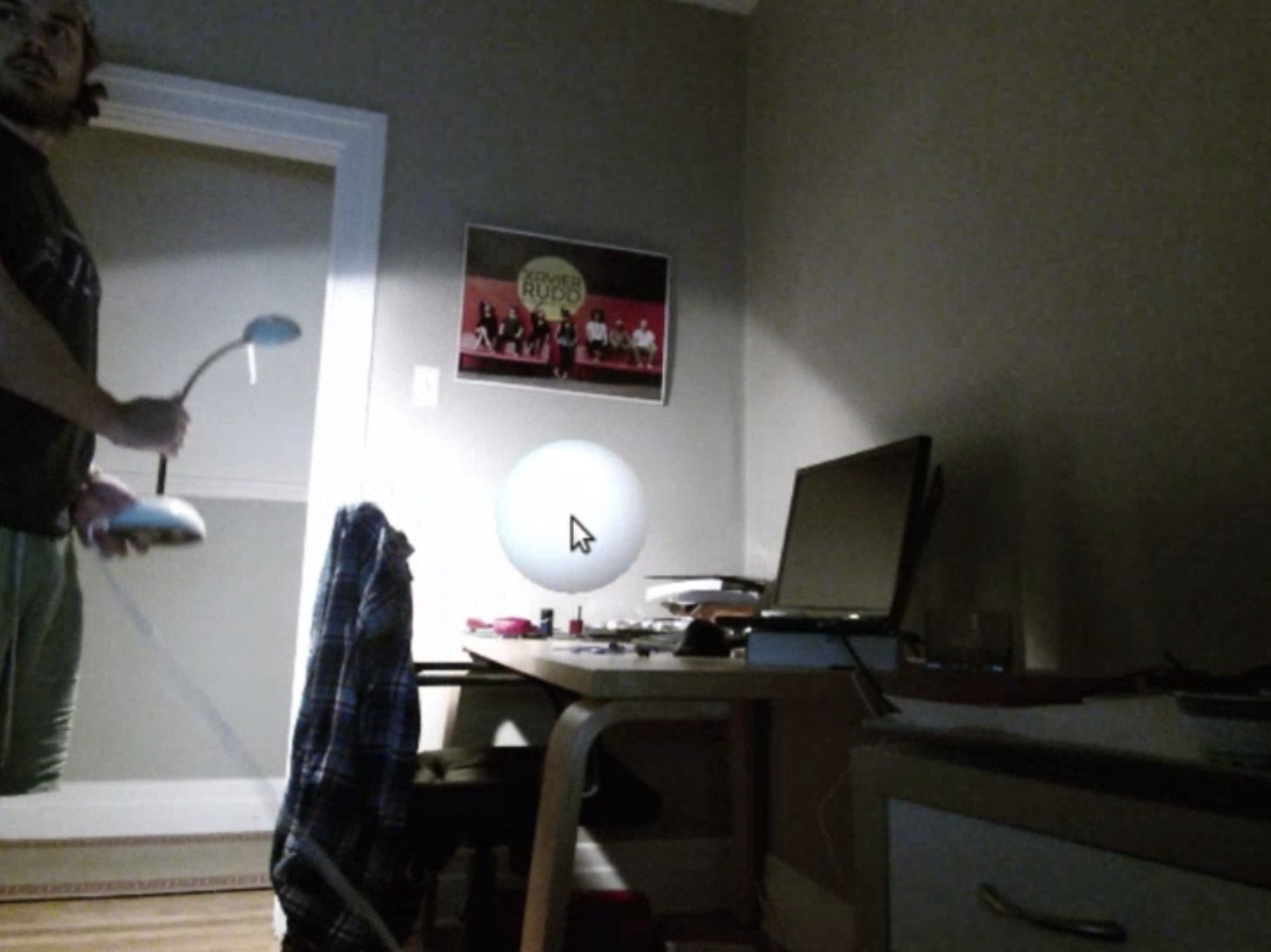} &
\includegraphics[width=.5\linewidth]{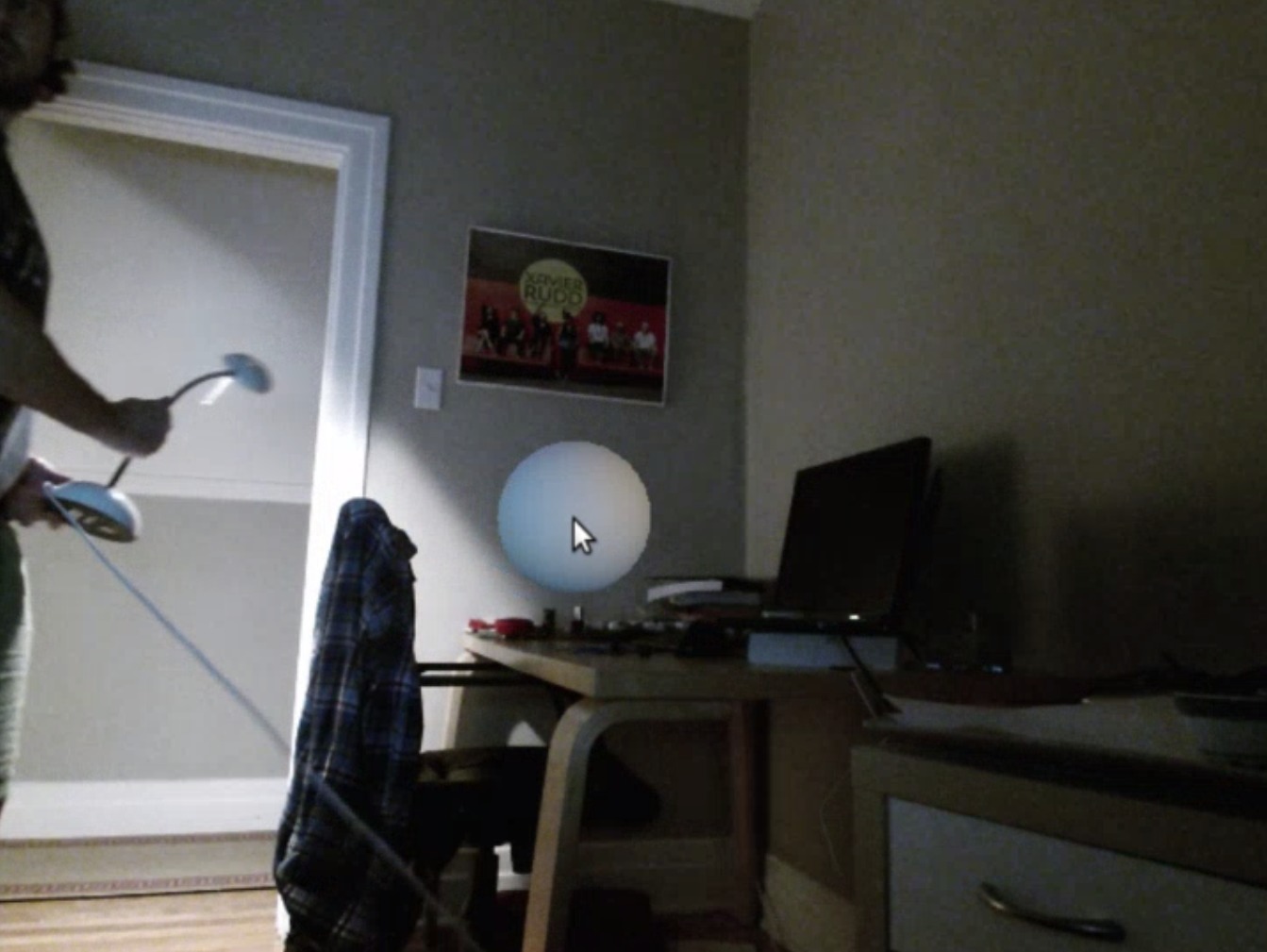} \\ 
\multicolumn{2}{c}{(b) Moving the light source} \\
\end{tabular}
\vspace{.25em}
\caption{Demonstration of real-time relighting applications. In (a), the object is moved through the scene and automatically relit at every frame. In (b), the object is static but a light source is moving around in the scene.}
\label{fig:demo}
\end{figure}

Our approach is particularly well suited for real-time applications, with an execution time of 20ms per image on an Nvidia 970M mobile GPU. We demonstrate our approach in live demonstrations recorded and available as a supplementary video to this paper, and show representative frames in fig.~\ref{fig:demo}. For these demonstrations, a virtual sphere is placed in the scene and illuminated with the SH lighting estimated by the network at every frame independently (no temporal consistency is enforced). The object is relit while the user clicks and drags it across the scene. We record the video feed using a Kinect V2: \emph{only the RGB image is given to the network for lighting estimation}. The depth frame is merely used to rescale the object. We demonstrate two scenarios. First, the camera is kept static, and the user drags the virtual object across the scene (fig.~\ref{fig:demo}-(a)). Second, both the camera and object are static, and the user moves a (real) light source around (fig.~\ref{fig:demo}-(b)). This shows that our approach adapts to strongly varying local lighting effects in real-time and stably, despite the lack of enforced temporal consistency.

\section{Conclusion and Future Work}

We present a real-time method, particularly suitable for AR, to predict local lighting for indoor scenes. As demonstrated via extensive evaluations on synthetic and real data, our method significantly outperforms previous work.

We envision a number of directions for future exploration. Our method is currently applied to single images. While we were surprised with the temporal stability of our results when applied to each frame of a video stream, re-training our network with temporal image information may increase the prediction accuracy. Video input also opens up the possibility of \emph{aggregating} scene information across frames to produce more accurate predictions. A future direction will be to explore different lighting representations to improve the angular frequency of our predictions leading to crisper shadows, and ultimately suitable reflection maps, for a seamless physically-based rendering pipeline and AR experience.

\section*{Acknowledgments}
This work was partially supported by the REPARTI Strategic Network and the NSERC/Creaform Industrial Research Chair on 3D Scanning: CREATION 3D. We gratefully acknowledge the support of Nvidia with the donation of the GPUs used for this research, as well as Adobe for generous gift funding.

{\small
\bibliographystyle{ieee}
\bibliography{refs}
}

\end{document}